\documentclass{article}

\PassOptionsToPackage{numbers, sort&compress}{natbib}

\usepackage[preprint]{neurips_2026}

\usepackage[utf8]{inputenc} % allow utf-8 input
\usepackage[T1]{fontenc}    % use 8-bit T1 fonts
\usepackage{hyperref}       % hyperlinks
\usepackage{url}            % simple URL typesetting
\usepackage{booktabs}       % professional-quality tables
\usepackage{amsfonts}       % blackboard math symbols
\usepackage{nicefrac}       % compact symbols for 1/2, etc.
\usepackage{microtype}      % microtypography
\usepackage{xcolor}         % colors
\usepackage{xspace}
\usepackage{amsmath}
\usepackage{graphicx}
\usepackage{subfigure}
\usepackage{enumitem}
\usepackage{wrapfig}
\usepackage{booktabs}
\usepackage{graphicx}
\usepackage{multirow}

\newcommand{\name}{TKCAM\xspace}
\newcommand{\dsname}{RealEstate10K-Cap\xspace}
\newcommand{\eg}{e.g.,\xspace}

\title{TKCAM: Text and Keyframe to Camera Trajectory Generation}

\author{%
\bfseries
Haozhe Yang\textsuperscript{1} \quad
Zhiyang Dou\textsuperscript{1,$\dagger$} \quad
Zekai Gu\textsuperscript{2} \quad
Cheng Lin\textsuperscript{3} \\
\bfseries
Wenping Wang\textsuperscript{4} \quad
Yuan Liu\textsuperscript{2} \quad
Taku Komura\textsuperscript{1,$\dagger$} \\[3pt]
\normalfont
\textsuperscript{1}The University of Hong Kong \quad
\textsuperscript{2}The Hong Kong University of Science and Technology \\
\textsuperscript{3}Macau University of Science and Technology \quad
\textsuperscript{4}Texas A\&M University \\[2pt]
\texttt{\{yhz2004, zhiyang0\}@connect.hku.hk} \quad
\texttt{taku@cs.hku.hk} \\
\texttt{zekai.gu@connect.ust.hk} \quad
\texttt{yuanly@ust.hk} \\
\texttt{chenglin@must.edu.mo} \quad
\texttt{wenping@tamu.edu}
}

\begin{document}

\maketitle

\begingroup
\renewcommand{\thefootnote}{\fnsymbol{footnote}}
\footnotetext[2]{Corresponding authors.}
\endgroup

\begin{abstract}
Generating high-quality and controllable camera motion is essential for AI-assisted cinematography, video synthesis, and 3D scene understanding. We introduce \textbf{\name}, a \textit{Text-} and \textit{Keyframe-}conditioned \textit{CAM}era-motion synthesis framework based on generative masked modeling. We represent camera dynamics using a 12-dimensional kinematic feature comprising position, velocity, and a continuous rotation representation and discretize them into hierarchical motion tokens via a Residual Vector Quantizer (RVQ). A two-stage masked transformer architecture then learns to reconstruct and refine these tokens, utilizing explicit self- and cross-attention modules for multimodal conditioning. A central feature of our framework is sparse visual keyframe conditioning: users can provide free-form text prompts together with RGB observations at selected timestamps, which provide temporally localized visual guidance for generating coherent in-between trajectories. Furthermore, to advance evaluation standards, we curate \textbf{\dsname}, a large-scale text-camera dataset, and establish a cross-domain benchmark with a Universal CLaTr Evaluator. Extensive experiments demonstrate that \name\ surpasses recent state-of-the-art baselines on Fréchet distance (FID), text-motion matching scores, and retrieval metrics (R@K), while
additional analyses evaluate temporal smoothness and cross-domain generalization. Code and the \dsname dataset are available at \url{https://github.com/linearalgebrayhz/TKCAM}.
\end{abstract}

\section{Introduction}

Camera motion determines how a scene is perceived, understood, and narrated. For modern vision–language systems, the ability to synthesize high-quality and controllable 6-DoF camera trajectories enables AI-assisted cinematography~\cite{jiang2020example, jiang2024cinematographic}, video synthesis~\cite{he2025cameractrl,yu2024viewcrafter,zheng2024cami2v,xu2024camco,kuang2024collaborative,li2025martian}, and spatial reasoning in 3D environments~\cite{yu2025hero, patel2025uniegomotion,xiong2025vision,bajcsy1988active}. Yet translating sparse, symbolic language into metrically consistent 3D motion remains challenging: language under-specifies geometry and timing; naïve regression drifts semantically; and existing datasets rarely couple reliable camera annotations with temporally grounded captions or standardized tools for evaluating text and camera-motion alignment.

Recent approaches~\cite{jiang2024cinematographic, courant2024exceptional, li2024director3d, zhang2025gendop} have modeled text-to-trajectory either autoregressively or via diffusion. While these models can produce smooth paths given prompts, they still face three practical limitations. \textit{First}, direct regression from text often ignores fine-grained visual semantics, leading to trajectories that miss spatial relations or temporal emphasis. \textit{Second}, existing control mechanisms commonly rely on explicit camera states, scene geometry, or other structured inputs, whereas users may instead wish to guide motion using sparse visual observations at selected timestamps. \textit{Third}, the modality gap between discrete text and continuous 3D motion, combined with the limited availability of semantically rich supervision, hinders efficient learning and generalization.

In this paper, we address these issues with \textbf{\name}, a text- \emph{and} keyframe-conditioned camera-motion generator built on \emph{generative masked modeling}. \name\ first discretizes continuous 3D camera trajectories into hierarchical motion tokens using a Residual Vector Quantizer (RVQ). To learn the multimodal correspondences, we train a base masked transformer that performs conditional demasking to reconstruct the masked motion tokens. Instead of complex token-level masking heuristics, we adopt a simple attention design within the transformer blocks: a motion-motion self-attention module to capture the temporal dynamics of the trajectory, followed by a motion-condition cross-attention module to directly inject the text and visual keyframe embeddings. A \emph{residual masked transformer} then further refines the higher RVQ quantization levels in a coarse-to-fine manner, enhancing the high-frequency fidelity of the camera motion. As a result, at inference, \name\ supports both text-only generation and \emph{animator-style keyframe conditioning}: users can provide free-form text prompts alongside sparse RGB keyframes (e.g., first-frame or start/end observations), which serve as temporally localized visual guidance while the model generates coherent in-between trajectories conditioned on both the text and visual observations.

To train \name, we curate \textbf{\dsname}, a large-scale camera motion dataset built upon RealEstate10K \cite{zhou2018stereo}. {\dsname} couples high-quality 3D camera trajectories with temporally grounded directorial captions. Specifically, the corpus comprises \textbf{25K} annotated sequences (\textbf{5 million} frames, approximately \textbf{50} hours in total) across diverse real estate environments. Furthermore, a significant bottleneck in current camera generation research is the absence of a standardized and cross-domain evaluation metric. To support evaluation across different camera-motion domains, we further construct a Universal CLaTr Evaluator by retraining the CLaTr contrastive objective \cite{courant2024exceptional} on a balanced mixture of indoor, cinematic, and character-centric trajectory data.

We conduct extensive experiments to validate that \name\ generates semantically faithful camera trajectories with smooth translational motion under both text-only and text-keyframe settings. On our mixed-domain benchmark, \name\ outperforms recent state-of-the-art baselines on Fréchet distance (FID), CLaTr-based text–motion matching score, and retrieval metrics (R@K). Furthermore, ablations support the benefits of our 12-dimensional kinematic representation, RVQ discretization, and the dual masked transformer architecture. Our main contributions can be summarized as follows:

\begin{itemize}[leftmargin=*,itemsep=1pt,parsep=0pt,topsep=2pt]
  
    \item We present \textbf{\name}, a hierarchical masked generative framework tailored to multimodal camera trajectory synthesis. Building on established RVQ-VAE and masked modeling techniques, \name introduces a camera-specific 12-dimensional kinematic representation together with decoupled motion self-attention and multimodal cross-attention for camera-motion generation.
    
      \item We introduce animator-style sparse-keyframe conditioning for text-driven camera generation. RGB observations at selected timestamps provide soft visual guidance through cross-attention, enabling bidirectional generation of coherent in-between trajectories without requiring explicit camera extrinsics as user input.
      
    \item We curate \textbf{\dsname}, a large-scale dataset of reconstructed camera trajectories paired with temporally grounded, scene-aware captions, and establish a cross-domain evaluation protocol with a \emph{Universal CLaTr Evaluator}. We further validate the framework through controlled baseline
    retraining, temporal-smoothness analysis, and human evaluation of the proposed evaluator. Among the compared methods, ours achieves the best FID, matching, and retrieval scores.
\end{itemize}

\section{Related Work}

\noindent
\textbf{Camera Trajectory Generation.}
Early efforts in camera trajectory generation relied on optimization-based motion planning \cite{blinn1988cinematography, Galvane2015automatecamera} or direct continuous regression, which frequently suffered from high-frequency spatial jitter and struggled with semantic alignment. To address these issues, recent frameworks employ VLM-based planners to orchestrate 3D coordinates from abstract text (e.g., ShotVerse~\cite{Yang2026shorVerse}), alongside generative paradigms spanning diffusion and discrete sequence modeling. CamSketch~\cite{Wei2025CamSketch} further explores interactive camera-trajectory authoring from user sketches and example-derived motion codes. One prominent line of research focuses on diffusion models: CCD~\cite{jiang2024cinematographic} introduced a text-guided diffusion model in character-centric coordinates, whose keyframe conditioning specifies explicit camera states; E.T.~\cite{courant2024exceptional} and Director3D~\cite{li2024director3d} subsequently leveraged Diffusion Transformers (DiT) to synthesize global trajectories conditioned on character dynamics and on text for subsequent 3D scene generation, respectively. However, despite their impressive generative quality, diffusion-based approaches inherently require dozens of iterative denoising steps, leading to high computational latency. Alternatively, auto-regressive discrete models (such as GenDoP~\cite{zhang2025gendop}) have also been explored to bypass diffusion latency; we discuss their limitations regarding unidirectional generation in our subsequent discussion on discrete sequence modeling.

\noindent
\textbf{Camera Trajectory Datasets.}
The evolution of camera trajectory datasets reflects a trade-off between geometric reliability and cinematographic expressiveness. Early datasets either relied on synthetic generation~\cite{jiang2024cinematographic} or leveraged structured captures of static scenes, such as MVImgNet2.0~\cite{han2024MvImgNet}, RealEstate10K~\cite{zhou2018stereo}, and DL3DV-10K~\cite{ling2023dl3dv}. While these datasets offer high-quality, calibrated trajectories, they rarely come with descriptions of shot intent. To address this, recent benchmarks like E.T.~\cite{courant2024exceptional} and DataDoP~\cite{zhang2025gendop} extract trajectories directly from feature films accompanied by rich textual descriptions. However, cinematic footage frequently contains artistic effects---such as motion blur, rapid cuts, and shallow depth-of-field---which can introduce severe artifacts during Structure-from-Motion (SfM) reconstruction, deviating from stable, natural navigation behaviors. Our curated \dsname\ corpus reaches a balance: it preserves the geometric precision of stable, real-world indoor captures while circumventing the reconstruction hazards of movie clips. DataDoP adopts a related two-stage strategy that derives motion descriptions from trajectories before language-model captioning; our contribution is the resulting indoor dataset with scene-grounded captions rather than this captioning strategy itself.

\noindent
\textbf{Discrete Generative Modeling for Sequences.}
To avoid the instability of continuous high-dimensional regression and the high latency of diffusion, another line of research formulates trajectory generation as a discrete sequence modeling task. Auto-regressive (AR) models, such as GenDoP~\cite{zhang2025gendop}, utilize tokenizers and transformers to predict camera movements token-by-token. While AR models excel at capturing discrete multimodal distributions, their strict sequential generation nature often leads to error accumulation over long sequences. More critically, the unidirectional (left-to-right) autoregressive paradigm limits their ability to condition on future observations, making animator-style tasks like filling the gap between a start and an end keyframe more difficult.

Building upon discrete paradigms, \name\ is designed to avoid both the iterative sampling of diffusion and the unidirectional generation of AR models. Following MoMask~\cite{guo2024momask}, we combine a 12-dimensional kinematic input, an RVQ tokenizer, and a masked generative framework for camera trajectory generation. Rather than generating frame-by-frame, our dual masked transformer predicts trajectories holistically across time. This enables parallel decoding and bidirectional keyframe conditioning via masked generation, with an explicit cross-attention mechanism for multimodal conditioning.

\section{Dataset}

\noindent
\textbf{Data Source and Preprocessing.} We construct our dataset upon the RealEstate10K collection~\cite{zhou2018stereo}, which provides high-quality SfM-reconstructed camera trajectories across diverse indoor scenes. To obtain clean and reliable motion data, we apply a series of rigorous preprocessing steps. We discard videos shorter than 4 seconds to ensure sufficient motion variation, and truncate continuous motions longer than 10 seconds to maintain temporal consistency. We further exclude near-static segments by filtering out trajectories where the cumulative camera displacement falls below a predefined threshold. After filtering, the curated set contains approximately 25K high-fidelity trajectory sequences, preserving a rich diversity of cinematographic indoor motion patterns.

\noindent
\textbf{Caption Generation.} A major challenge in building text-motion datasets is the tendency of Vision-Language Models (VLMs) to hallucinate 3D camera dynamics. To keep motion descriptions consistent with the trajectories while adding scene detail, we leverage a \textbf{two-stage captioning pipeline} that couples deterministic trajectory analysis with generative visual grounding. 

In the first stage, we derive a strict geometric motion prior directly from the relative camera poses. We integrate local camera velocities while suppressing minor pose-estimation jitter (threshold $<0.005$). We then apply rule-based kinematic thresholds to classify translational (e.g., track, dolly) and rotational (e.g., pan, tilt) movements based on the OpenGL convention. Notably, simultaneous lateral tracking and yaw are merged and labeled as cinematic \textit{arcs}. This step yields a rule-derived motion description (\eg \textit{``camera dollies forward, then pans left''}).

In the second stage, we employ Qwen3-VL-4B-Instruct~\cite{qwen2025qwen3vl, yang2025qwen3} to ground this motion prior within the visual context. For each sequence, we uniformly sample up to 32 video frames. The VLM is strictly prompted to treat the deterministic geometric prior as an immutable constraint (\textit{absolute motion}), while enriching it with salient scene elements, spatial layouts, and cinematic pace. Generation is performed using stochastic decoding (temperature $0.7$, top-$p$ $0.9$, max new tokens $128$). If visual frames are unavailable, the caption falls back to the rule-derived description. The full prompt template is provided in Appendix~\ref{sec:data_pipeline}.

\begin{figure}[t]
\vspace{-7mm}
  \centering
  \includegraphics[width=0.8\textwidth]{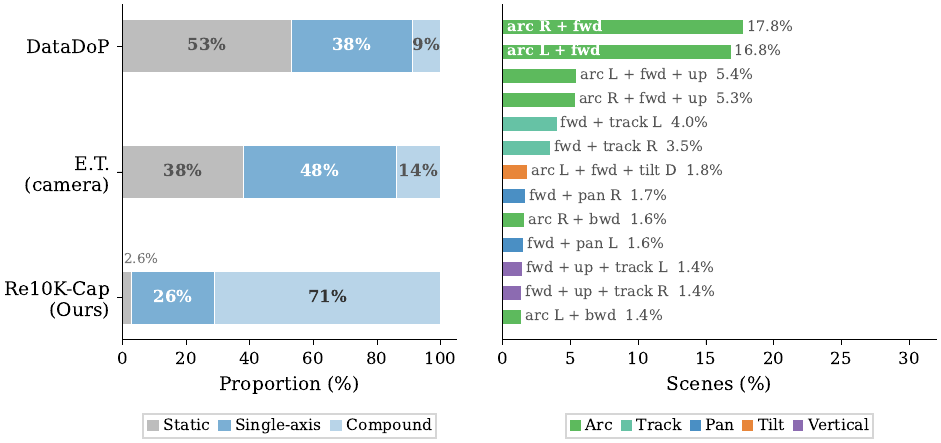}
\vspace{-2mm}
  \caption{Motion distribution comparison across camera trajectory datasets. 
  \textbf{Left:} Coarse-grained breakdown into static, single-axis, and 
  compound motion (DataDoP and E.T.\ values estimated from their published 
  figures). \textbf{Right:} Top-13 motion combinations in \dsname; the remaining 
  35.1\% is distributed across 50+ other combination patterns.}

  \label{fig:motion_distribution}
  \vspace{-6mm}
\end{figure}

\noindent
\textbf{Dataset Statistics.} Our dataset comprises 25K high-fidelity trajectory sequences sampled from diverse indoor environments, yielding approximately \textbf{5 million} frames and \textbf{50} hours of continuous footage. Each trajectory spans an average of \textbf{201.7} frames ($\sim$\textbf{6.72} seconds at 30 FPS), capturing smooth, uninterrupted camera navigation through complex real-world spaces. Table~\ref{tab:dataset_comparison} provides a comprehensive quantitative comparison with existing camera trajectory datasets.

\begin{wraptable}{r}{0.65\columnwidth}
\vspace{-6mm}
    \caption{Comparison with existing camera trajectory datasets. Our dataset focuses on indoor environments with scene-aware captions.}
    \label{tab:dataset_comparison}
    \centering
    \vspace{2mm}
    \footnotesize
    \setlength{\tabcolsep}{2.5pt}
    \resizebox{\linewidth}{!}{
    \begin{tabular}{@{}lccccc@{}}
        \toprule
        \textbf{Dataset} & \textbf{\#Samples} & \textbf{Domain} & \textbf{Caption} & \textbf{Vocab} & \textbf{Avg. Length (s)} \\
        \midrule
        RealEstate10K~\cite{zhou2018stereo} & 79K & Indoor & \texttimes & - & - \\
        CCD~\cite{jiang2024cinematographic} & 25K & Synthetic & Synthetic & 48 & 7.2 \\
        E.T.~\cite{courant2024exceptional} & 115K & Film & Camera-Char & 1.7K & 3.8 \\
        DataDoP~\cite{zhang2025gendop} & 29K & Film & Directorial & 8.7K & 14.4 \\
        \midrule
        \textbf{Ours} & \textbf{25K} & \textbf{Indoor} & \textbf{Scene-aware} & \textbf{3.8K} & \textbf{6.72} \\
        \bottomrule
    \end{tabular}
    }

    \vspace{-1em}
\end{wraptable}
% \subsection{Comparison with Existing Datasets}
\noindent
\textbf{Comparison with Existing Datasets.} Prior camera trajectory datasets mainly rely on synthetic environments or film clips. E.T.~\cite{courant2024exceptional} and DataDoP~\cite{zhang2025gendop}, for example, use feature films with stylized cinematography, but such footage often contains motion blur, rapid cuts, and shallow depth-of-field, causing SfM artifacts and deviating from stable, natural camera navigation. As shown in Figure~\ref{fig:motion_distribution}, film-derived datasets also suffer from motion imbalance: near-static shots comprise 53\% of DataDoP and 38\% of E.T., whereas our filtering reduces static content to 2.6\%. In contrast, 71\% of \dsname{} contains compound motion, compared to 9\% in DataDoP and 14\% in E.T. The top-13 motion patterns cover diverse behaviors such as forward arcs, dolly-track combinations, and multi-axis motion, with the remaining 35.1\% spread across 50+ patterns. These statistics characterize the motion distribution of RealEstate10K-Cap but do not by themselves establish downstream dataset utility. We therefore additionally conduct a controlled comparison by training the same \name\ architecture on equally sized subsets of RealEstate10K-Cap and DataDoP; the full results are provided in Appendix~\ref{sec:dataset_controlled}. Unlike film-based datasets, RealEstate10K-Cap is built from stable real-estate tours, yielding reliable 3D reconstructions, while its two-stage captioning pipeline provides trajectory-grounded motion descriptions together with rich spatial context.

\section{Method}

\begin{figure}[t]
\vspace{-6mm}
  \centering
  \includegraphics[width=\textwidth]{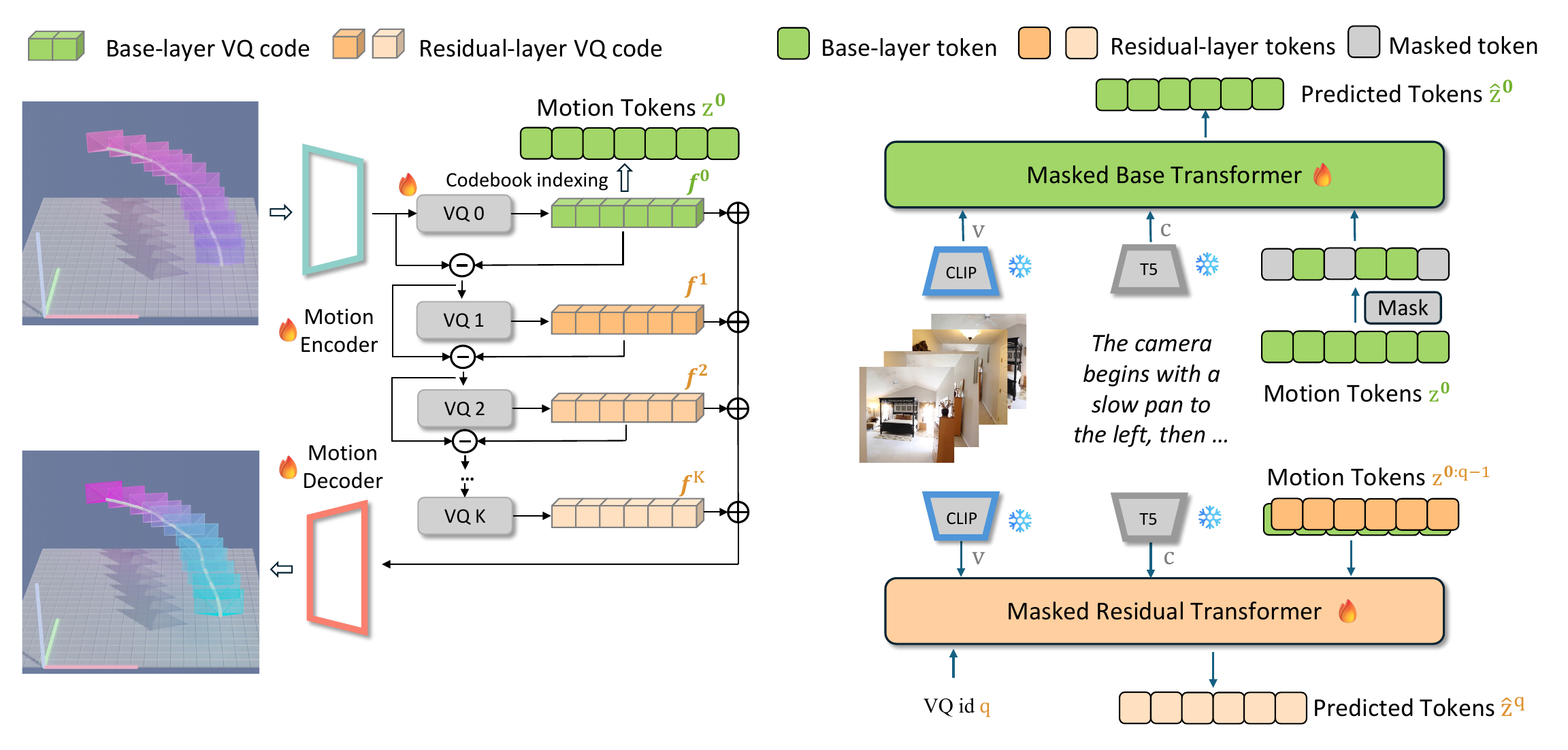}

\vspace{-3mm}

  \caption{Overview of our proposed pipeline. 
  Given per-frame camera poses $\mathcal{M}$ as trajectory, a text description $\mathcal{T}$ and sparse keyframe images $\mathcal{V}$, 
  we first encode the continuous 12-dimensional camera trajectory into latent features and discretize them with a multi-level Residual Vector Quantizer (RVQ), producing a hierarchy of base-layer (green) and residual-layer (orange) motion tokens. 
  During training, a masked base transformer predicts masked base-layer tokens conditioned on the text embedding from a frozen T5 encoder and visual tokens extracted by a frozen CLIP encoder, while a masked residual transformer refines higher-level residual tokens conditioned on the previously predicted tokens and the same multimodal context. 
  Finally, a motion decoder reconstructs smooth and temporally coherent camera trajectories from the predicted hierarchical motion tokens.}
  \label{fig:main_pipeline}
  \vspace{-5mm}

\end{figure}

% \subsection{Overview}

Given a text description $\mathcal{T}$ and a sparse set of keyframe images $\mathcal{V} = \{v_t\}_{t=1}^{T_k}$, our goal is to generate a temporally coherent and semantically consistent camera trajectory $\mathcal{M} = \{m_t\}_{t=1}^{T}$ that is conditioned on these keyframes and aligned with the described scene dynamics.

Our overall pipeline is illustrated in Fig.~\ref{fig:main_pipeline}. We discretize the continuous camera trajectories into hierarchical motion tokens using a Residual Vector Quantizer (Sec.~\ref{sec:rvq}). A masked generative modeling framework is then adopted to learn multimodal correspondences between text, RGB keyframes, and motion tokens (Sec.~\ref{sec:masked}), where base and residual masked transformers collaboratively predict and refine hierarchical motion tokens for coherent trajectory generation.

\subsection{Motion Representation and Quantization}
\label{sec:rvq}

To ensure mathematical continuity and mitigate the instability of direct high-dimensional regression, we formulate the camera trajectory at frame $t$ as a 12-dimensional kinematic vector $m_t = [p_t, v_t, r_t]$. This incorporates normalized absolute coordinates $p_t$, per-frame linear velocities $v_t$ (acting as an implicit temporal regularization to smooth trajectories), and a continuous 6D rotation representation $r_t$ to avoid singularities \cite{zhou2019continuity}. To better capture the modes of cinematographic dynamics, we map this continuous 12D sequence into discrete tokens~\cite{Zhang2023t2mgpt} using Residual Vector Quantization (RVQ)~\cite{zeghidour2021soundstream, guo2024momask}, transforming the geometric motion modeling into a tractable discrete sequence generation problem.

\noindent\textbf{Hierarchical Motion Tokens.} 
The continuous kinematic camera trajectories $m_t = [p_t, v_t, r_t]$ are first passed through a lightweight encoder composed of 1D convolutions and residual connections. This maps the motion sequence into latent embeddings $z \in \mathbb{R}^d$. These embeddings are then quantized by a multi-level RVQ module equipped with $K$ codebooks. The quantized representation $\hat{z}$ is obtained by summing the selected codewords across valid quantization levels: $\hat{z} = \sum_{k=1}^{K'} f_{q^{(k)}}$ where $f_{q^{(k)}}$ denotes the selected codeword from the $k$-th codebook. To encourage robust representation learning, we employ \textit{quantizer dropout}~\cite{zeghidour2021soundstream} during training, randomly truncating the number of active quantizers $K' \le K$ with a probability $p=0.2$. This hierarchical discrete design provides two main benefits: first, it transforms continuous regression into a stable token prediction task, improving semantic coherence; second, lower quantization levels capture coarse trajectory patterns (e.g., global direction), while higher levels progressively refine high-frequency motion details.

\noindent\textbf{Training Objective.}
To optimize the RVQ autoencoder, we employ a multi-component objective tailored to 12D kinematic properties. The codebooks are updated via Exponential Moving Average (EMA)~\cite{vandenoord2017vqvae, guo2024momask}, while the encoder is optimized by the commitment loss~\cite{vandenoord2017vqvae}: $\mathcal{L}_{c} = \frac{1}{K'} \sum_{k=1}^{K'} \| z^{(k)} - \mathrm{sg}[f_{q^{(k)}}] \|_2^2$, where $z^{(k)}$ is the residual feature at level $k$, $f_{q^{(k)}}$ is its selected codeword as defined above, and $\mathrm{sg}[\cdot]$ is the stop-gradient operator.

For trajectory reconstruction, let $\ell(x, \hat{x})$ denote the Smooth L1 (Huber) loss. The RVQ objective $\mathcal{L}_{\text{RVQ}}$ combines a global reconstruction term $\mathcal{L}_{\text{rec}} = \ell(m_t, \hat{m}_t)$ with an explicit kinematic loss $\mathcal{L}_{\text{exp}}$ and commitment weight $\beta$:
\begin{equation}
    \mathcal{L}_{\text{RVQ}} = \mathcal{L}_{\text{rec}} + \lambda_{\text{exp}} \mathcal{L}_{\text{exp}} + \beta \mathcal{L}_{c}
\end{equation}
Specifically, $\mathcal{L}_{\text{exp}}$ independently supervises translation-velocity and 6D rotation components to ensure structural integrity: $\mathcal{L}_{\text{exp}} = \ell([p_t, v_t], [\hat{p}_t, \hat{v}_t]) + \ell(r_t, \hat{r}_t) + \lambda_{\text{orth}} \mathcal{L}_{\text{orth}}$. The term $\mathcal{L}_{\text{orth}}$ enforces the orthogonality of the predicted 6D rotation vectors $\hat{r}_1, \hat{r}_2 \in \mathbb{R}^3$:
\begin{equation}
    \mathcal{L}_{\text{orth}} = \mathbb{E} \left[ (\|\hat{r}_1\|_2 - 1)^2 + (\|\hat{r}_2\|_2 - 1)^2 + (\hat{r}_1^\top \hat{r}_2)^2 \right]
\end{equation}
Following MoMask \cite{guo2024momask}, we set $\lambda_{\text{exp}} = 0.5$ and $\beta = 0.02$. $\lambda_{\text{orth}}$ is fixed at $0.1$. 

\subsection{Masked Generative Modeling}
\label{sec:masked}

Our framework adopts a two-stage masked generative modeling paradigm \cite{guo2024momask}. Following RVQ discretization, the continuous trajectory is mapped to a discrete latent space $Z \in \{0, \dots, V-1\}^{B \times S \times K}$, where $V$ is the codebook size, $S$ is the sequence length, and $K=4$ is the number of quantization levels. We train two separate networks: a base masked transformer to model the primary structural motion $Z^{(0)}$, and a residual masked transformer to predict fine-grained details $Z^{(1:K-1)}$.

\noindent\textbf{Base Masked Transformer.} The base transformer $p_\theta$ reconstructs the first-level motion tokens $Z^{(0)}$ from a corrupted observation. During training, a subset of valid tokens is replaced with a \texttt{[MASK]} token following a randomized cosine schedule \cite{Chang2022maskgit, guo2024momask}. To inject multimodal conditioning, we extract semantic embeddings $c$ via a frozen T5 encoder~\cite{raffel2020t5}, and visual tokens $v$ from sparse RGB keyframes via a frozen CLIP encoder~\cite{radford2021learning}. Crucially, to resolve spatial-temporal ambiguity (e.g., distinguishing start from end keyframes), we explicitly inject a learned temporal positional embedding into each visual token corresponding to its VQ-aligned timestamp. These visual features provide temporally localized conditioning and do not directly specify or clamp metric camera poses.

Rather than heuristically adding visual tokens to motion tokens, our architecture utilizes decoupled attention \cite{aksan2021spatiotemporaltransformer, alayrac2022flamingo}. Each transformer block sequentially executes motion-motion self-attention over the masked sequence $\tilde{Z}^{(0)}$ to capture internal kinematic dynamics, followed by motion-condition cross-attention querying the concatenated multimodal condition $[c; v]$. This ordering ensures motion tokens establish temporal coherence before attending to multimodal conditions. To enable Classifier-Free Guidance \cite{Ho2022CFG}, we independently drop $c$ and $v$ with a probability of $p=0.2$ during training. Finally, the model is optimized using a standard cross-entropy loss applied exclusively to the masked positions, seamlessly avoiding the need for complex token-level heuristics.

\noindent\textbf{Residual Masked Transformer.} To generate cinematographically plausible high-frequency details (e.g., subtle camera jitters), the residual masked transformer $p_\phi$ refines motion tokens level-by-level in a fully parallel manner. For an active residual layer $q \in \{1, \dots, K-1\}$ (sampled per batch via an inverted cosine schedule to emphasize lower levels), it predicts the residual tokens conditioned on the cumulative historical embeddings of all previous levels $H^{(q)} = \sum_{j=0}^{q-1} E(Z^{(j)})$, alongside the identical multimodal conditions $c$ and $v$. The network shares weights across all quantization levels, utilizing $q$ as an additional level-embedding. It is optimized by minimizing a conditional masked cross-entropy loss over the temporal dimension for the active layer. This coarse-to-fine framework allows \name\ to establish a stable macroscopic trajectory structure before progressively enriching it with microscopic physical realism. Complete objective formulation can be found in Appendix \ref{sec:appendix_objectives}.

% \subsection{Inference}
During inference, given a target trajectory length, we extract text and visual embeddings $c,v$ via T5 and CLIP encoders and initialize the corresponding motion-token sequence as fully masked, including positions temporally aligned with the visual keyframes. The base transformer reconstructs the motion tokens using parallel decoding and Classifier-Free Guidance (CFG) to establish the coarse structure $\hat{Z}^{(0)}$. Subsequently, the residual transformer refines the trajectory across all quantization levels in a feed-forward manner to predict $\hat{Z}^{(1:K-1)}$. The final sequence is passed to the RVQ decoder to produce the continuous trajectory:
\begin{equation}
\hat{m}_t = \text{Decoder}_{\text{RVQ}}\big(\hat{Z}^{(0)}, \dots, \hat{Z}^{(K-1)}\big).
\end{equation}
This mechanism supports both text-only generation and animator-style keyframe conditioning. 

\section{Experiment}

\noindent
\textbf{Datasets and Setup.}
\name\ is trained exclusively on RealEstate10K-Cap and evaluated zero-shot on the E.T. and DataDoP test subsets, demonstrating cross-domain generalization without any in-domain supervision on cinematic or character-centric data. All trajectories are normalized to first-frame-relative coordinates to ensure spatial comparability. We compare with CCD~\cite{jiang2024cinematographic}, E.T.~\cite{courant2024exceptional}, Director3D~\cite{li2024director3d}, and GenDoP~\cite{zhang2025gendop}. Following~\cite{zhang2025gendop}, CCD and E.T. receive zeroed character-motion tokens at inference.
 Since GenDoP requires RGB-D input, we estimate depth with Depth Anything V2~\cite{v2024depthanythingv2} and fine-tune GenDoP on RealEstate10K-Cap. For a controlled comparison, we additionally fine-tune/retrain CCD and E.T. on RealEstate10K-Cap under the same evaluation protocol, and report both released pretrained and matched-training results where available.

\name\ requires a target trajectory length at inference. For benchmark evaluation, we use the per-sample ground-truth trajectory length for all methods, with every trajectory evaluated at the same target duration. Thus, the target generation horizon is assumed to be known during evaluation. Baseline training and trajectory-length handling are detailed in Appendix~\ref{sec:baseline_protocol} and~\ref{sec:trajectory_length}.

We use the CLaTr framework~\cite{courant2024exceptional}, training an evaluator on 15K validation samples balanced across RealEstate10K-Cap, E.T., and DataDoP, all disjoint from model training data. Results are reported on a separate 1.5K-sample test set. Metrics include Matching Score and retrieval ($R@1,R@3,R@10$) for text-motion alignment, FID~\cite{heusel2017fid}, computed on Universal CLaTr features, for distributional realism, and $\Delta$Div for diversity alignment.

To validate the Universal CLaTr evaluator against human judgment, we further conduct a pilot blinded human study over 90 caption-trajectory pairs from four methods and the ground-truth reference. Universal CLaTr shows a positive correlation with mean human judgments (Spearman $\rho=0.312$, 95\% CI $[0.089, 0.492]$). Full study details are provided in Appendix~\ref{sec:human_study}.

\noindent
\textbf{Implementation Details.}
All experiments run on a single NVIDIA A100 80GB GPU. Our RVQ-VAE uses 256-code codebooks and 4 quantization stages. Both base and residual transformers have 4 layers, 6 heads, and hidden size 384, with T5 and CLIP ViT-B/32 as multimodal encoders. Non-autoregressive decoding yields about 30 trajectories/s. Full hyperparameters are in Appendix~\ref{sec:hyperparameters}.

\subsection{Main Results}

\begin{table}[t]
\vspace{-5mm}
  \centering
  \caption{Quantitative evaluation of text-to-camera trajectory generation on the mixed benchmark. Metrics include Fréchet distance computed on Universal CLaTr features (FID), text-trajectory Matching Score (mean cosine similarity in $[-1,1]$), and Retrieval metrics (R@$1$, R@$3$, R@$10$, in \%) computed over the full evaluation set of each track. $\Delta$Div represents the absolute difference between the generated trajectory diversity and the reference set diversity (closer to 0 is better). ``Fine-tuned'' denotes models fine-tuned/retrained on RealEstate10K-Cap. \textbf{Bold} indicates the best performance within each track.}
  \label{tab:main_exp_results}
\resizebox{\linewidth}{!}{
  \begin{tabular}{lcccccc}
    \toprule
    \textbf{Method} & \textbf{FID} $\downarrow$ & \textbf{Matching} $\uparrow$ & \textbf{R@1} $\uparrow$ & \textbf{R@3} $\uparrow$ & \textbf{R@10} $\uparrow$ & $\Delta$\textbf{Div} $\downarrow$ \\
    \midrule
    \multicolumn{7}{c}{\textit{Track A: Text-Conditioned (1500 samples)}} \\
    \midrule
    CCD \cite{jiang2024cinematographic}
        & 0.688 & 0.003 & 0.00 & 0.20 & 0.93 & 0.140 \\
    CCD (fine-tuned) \cite{jiang2024cinematographic}
        & 0.577 & 0.037 & 0.27 & 0.80 & 2.20 & 0.109 \\
    E.T. \cite{courant2024exceptional}
        & 1.167 & -0.013 & 0.00 & 0.27 & 1.00 & 0.502 \\
    E.T. (fine-tuned) \cite{courant2024exceptional}
        & 0.577 & 0.040 & 0.20 & 0.80 & 2.20 & 0.095 \\
    Director3D \cite{li2024director3d}
        & 0.849 & -0.002 & 0.00 & 0.07 & 0.53 & 0.186 \\
    GenDoP (Text only) \cite{zhang2025gendop}
        & 0.760 & 0.029 & 0.20 & 0.40 & 2.13 & 0.138 \\
    GenDoP (Text only, fine-tuned) \cite{zhang2025gendop}
        & 0.507 & 0.069 & 0.87 & 2.33 & 5.40 & 0.076 \\
    \textbf{\name (Text Only)}
        & \textbf{0.476} & \textbf{0.094} & \textbf{1.00} & \textbf{2.87} & \textbf{8.40} & \textbf{0.064} \\
    \midrule
    \multicolumn{7}{c}{\textit{Track B: Vision- \& Text-Conditioned (Visual Keyframe Conditioning, 994 samples)}} \\
    \midrule
    GenDoP (RGBD) \cite{zhang2025gendop}
        & 0.835 & 0.043 & 0.60 & 1.21 & 3.92 & 0.206 \\
    GenDoP (RGBD, fine-tuned) \cite{zhang2025gendop}
        & 0.566 & 0.098 & 1.01 & 3.62 & 8.25 & 0.104 \\
    \textbf{\name (First-Frame)}
        & \textbf{0.461} & 0.120 & 1.61 & 4.63 & 12.27 & 0.079 \\
    \textbf{\name (Sparse Keyframes)}
        & 0.462 & \textbf{0.129} & \textbf{2.82} & \textbf{5.43} & \textbf{15.19} & \textbf{0.071} \\
    \bottomrule
  \end{tabular}
  }
\end{table}

\begin{figure}[!t]
\vspace{-5mm}
    \centering
    \includegraphics[width=\textwidth]{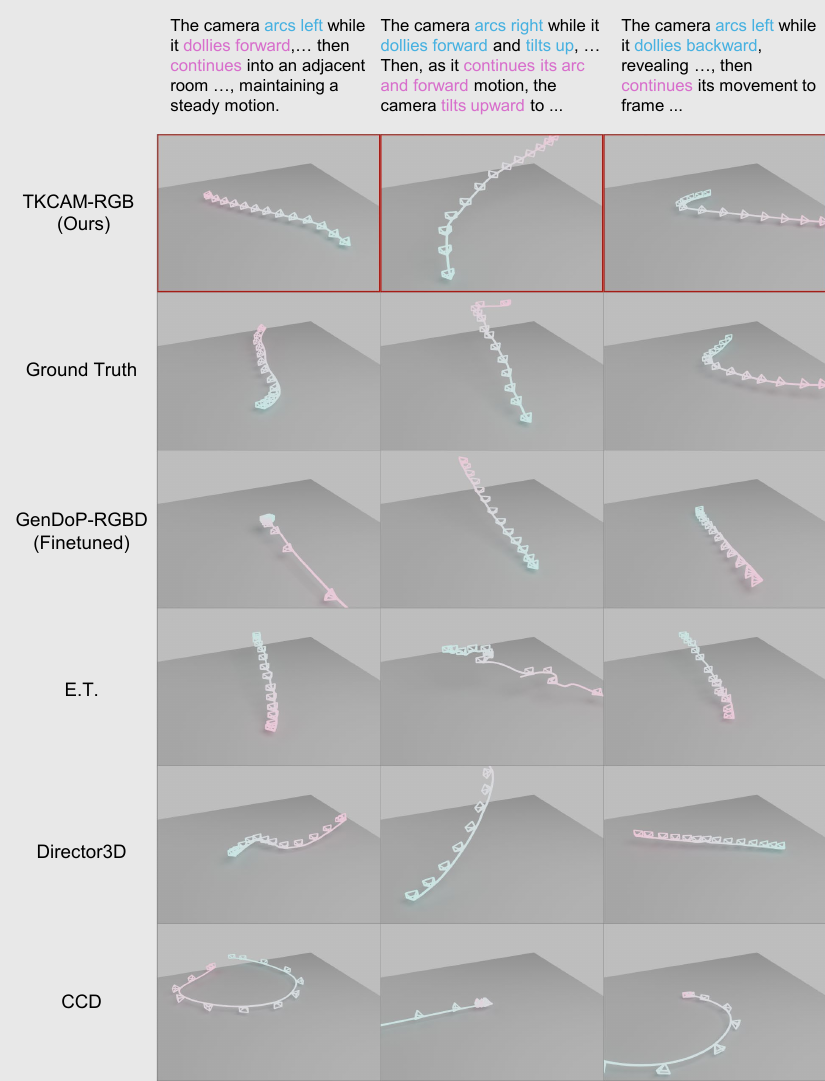}
    \vspace{-0.6cm} 
    \caption{Qualitative comparison of generations. While baselines struggle with motion magnitude, directional alignment, or trajectory quality (e.g., CCD's circular loops), \name produces semantically faithful paths that capture complex compound motions.}
    \label{fig:baseline_comparison}
    \vspace{-6mm}
\end{figure}

\noindent\textbf{Text-to-Camera Generation.}
In Track A (Table~\ref{tab:main_exp_results}), \name\ (Text Only) achieves the best results among the compared methods across all six metrics. It attains an R@10 of \textbf{8.40\%} and a Matching Score of \textbf{0.094}, compared with 5.40\% and 0.069 for the strongest text-only baseline, fine-tuned GenDoP. It also achieves the lowest FID of \textbf{0.476} and diversity gap of \textbf{0.064}, indicating closer agreement with the reference trajectory distribution in the evaluator's feature space. The examples in Figure~\ref{fig:baseline_comparison} illustrate \name's ability to capture compound camera motions with correct directional semantics and smooth phase transitions. In these examples, the baselines exhibit reduced motion magnitude, directional misalignment, near-static motion, or circular loops that do not follow the prompts.

\begin{figure}[t] 
    \centering
    \includegraphics[width=0.95\textwidth]{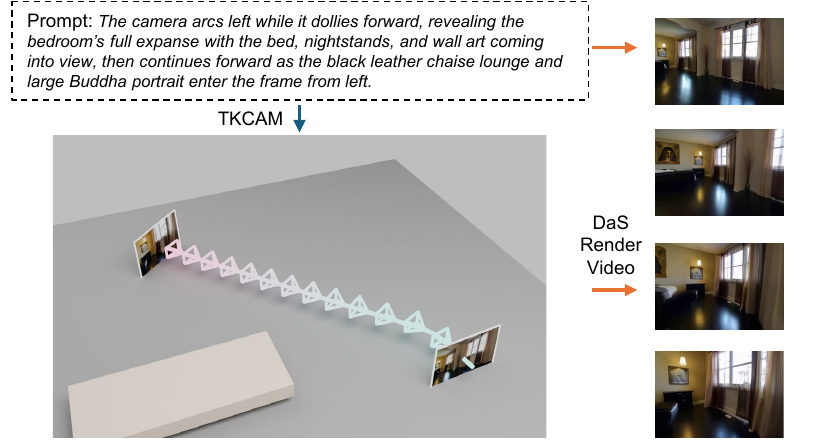}
    \vspace{-0.2cm}
    \caption{\name\ as a geometric motion prior for downstream video generation tasks.}
    \label{fig:video_consistency}
    \vspace{-4mm}
\end{figure}

\noindent\textbf{Keyframe-Conditioned Generation and Video Synthesis.}
In Track B, both visual-conditioning configurations of \name\ outperform the compared RGBD-conditioned GenDoP baselines across all six metrics. Using RGB input without explicit depth conditioning, \name\ (First-Frame) achieves an FID of \textbf{0.461} and an R@10 of 12.27\%, compared with 0.566 and 8.25\% for fine-tuned GenDoP (RGBD). The Sparse Keyframes configuration achieves the highest Matching Score of \textbf{0.129} and R@10 of \textbf{15.19\%} within this track, while retaining a nearly identical FID of 0.462. We further demonstrate this animator-style conditioning in a downstream video synthesis pipeline (Figure~\ref{fig:video_consistency}). Given first- and last-frame observations alongside a directorial prompt, \name\ generates a temporally coherent trajectory that serves as a geometric motion prior for Diffusion-as-Shader~\cite{Gu2025das}. The resulting video exhibits stable camera motion consistent with the described motion, suggesting that the generated trajectories can serve as useful geometric priors for downstream video synthesis.

\noindent\textbf{Temporal Smoothness.}
We further evaluate translational acceleration/jerk and angular velocity/acceleration/jerk. \name\ with sparse keyframes closely matches the ground-truth translational dynamics, while rotational acceleration and jerk remain less well matched than GenDoP-RGBD, indicating rotational smoothness as a remaining limitation. Full derivative statistics and distributional comparisons are provided in Appendix~\ref{sec:smoothness_analysis}.

\subsection{Ablation Studies}
\label{sec:ablation}

We validate our design choices on the mixed-domain benchmark using unified training configurations. The 12D-representation and fusion ablations use the Sparse Keyframes setting (Track B), while the RVQ-level ablation in Table~\ref{tab:ablation_rvq_levels} uses the Text Only setting (Track A).
% TODO(authors): add the promised ablations (temporal keyframe embedding, cosine vs. uniform mask schedule; CLIP features and bidirectional generation if available) as a table, ideally in the appendix with a pointer here.

\noindent\textbf{12D Representation.}
Replacing our 12D formulation with a 9D variant (no velocity) degrades all metrics: FID 0.503$\to$0.593, Matching 0.129$\to$0.106, R@10 12.67\%$\to$10.66\%, and $\Delta$Div 0.070$\to$0.074, consistent with explicit velocity acting as an implicit temporal regularizer.

\noindent\textbf{Multimodal Fusion.}
Our decoupled cross-attention outperforms a \textit{Prefix Concatenation} baseline on both realism (FID 0.503 vs.\ 0.580) and alignment (R@10 12.67\% vs.\ 12.07\%, Matching 0.129 vs.\ 0.119), supporting the benefit of cross-attention for fine-grained temporal grounding.

\begin{wraptable}{r}{0.5\textwidth}
  \centering
  \vspace{-6mm}
  \footnotesize
  \setlength{\tabcolsep}{3pt}
  \caption{Ablation on hierarchical RVQ levels.}
  \label{tab:ablation_rvq_levels}
  \resizebox{\linewidth}{!}{
  \begin{tabular}{lcccccc}
    \toprule
    \textbf{RVQ Stages} & \textbf{FID}$\downarrow$ & \textbf{Match.}$\uparrow$ &
    \textbf{R@1}$\uparrow$ & \textbf{R@3}$\uparrow$ & \textbf{R@10}$\uparrow$ &
    $\Delta$\textbf{Div}$\downarrow$ \\
    \midrule
    L1 (Base Only)   & 0.582 & 0.094 & \textbf{1.40} & 4.07 & 9.33 & 0.081 \\
    L1 + L2          & \textbf{0.522} & \textbf{0.102} & 0.93 & 4.07 & 9.20 & 0.068 \\
    L1 + L2 + L3     & 0.525 & 0.099 & 1.33 & \textbf{4.27} & 8.73 & \textbf{0.063} \\
    \textbf{Full (L1--L4)} & 0.529 & 0.100 & 1.13 & 3.80 & \textbf{9.60} & 0.065 \\
    \bottomrule
  \end{tabular}}
\end{wraptable}

\noindent\textbf{Hierarchical Refinement.}
Table~\ref{tab:ablation_rvq_levels} reveals a trade-off across quantization depths: the base level alone (L1) attains the highest R@1 but the largest diversity gap, while adding residual levels lowers FID and the diversity gap at some cost to exact-match retrieval. The full model (L1--L4) achieves the best R@10 while maintaining competitive FID and diversity.

\noindent\textbf{Sparse Keyframe Conditioning.}
On the same 994-sample Track B evaluation set, adding sparse keyframes beyond the first frame improves the Matching Score from 0.120 to 0.129 and R@1/R@3/R@10 from 1.61/4.63/12.27\% to 2.82/5.43/15.19\%, respectively. The diversity gap also decreases from 0.079 to 0.071, while FID remains nearly unchanged (0.461 vs.\ 0.462). These results support the benefit of additional temporally localized visual guidance for text--trajectory alignment. As discussed in Sec.~\ref{sec:masked}, RGB keyframes provide soft visual guidance rather than explicit metric camera-pose constraints.

\section{Conclusion}

In this paper, we presented \name, a text- and keyframe-conditioned framework for camera trajectory generation. \name\ represents camera motion as 12-dimensional kinematic sequences and discretizes them with RVQ-VAE, reducing the instability of continuous regression. Its two-stage masked transformer enables multimodal fusion, high-frequency refinement, and animator-style keyframe conditioning. Experiments with our Universal CLaTr Evaluator demonstrate strong distributional realism and cross-modal alignment.

\noindent\textbf{Limitations and Future Work.}
\name\ has several limitations. First, the residual transformer is trained with ground-truth base tokens but uses predicted tokens at inference, which may introduce exposure bias. Second, training data is mainly indoor real-estate footage, leaving highly dynamic or stylized cinematography underexplored. Third, rotational smoothness remains weaker than translational smoothness, and inference requires a predefined target generation horizon. Finally, \name\ currently focuses on single-camera extrinsic motion; incorporating explicit scene geometry, camera intrinsics, and multi-camera coordination is a promising future direction.

\section*{Acknowledgments}

We thank Huaijin Pi, Boao Zhan, and Liangyu Zhang for valuable discussions. We are also grateful to the members of the HKU CGVU Lab for their helpful discussions and support, as well as for providing computational resources. Special thanks to Hanna Li for her personal support and encouragement.

\paragraph{Funding and competing interests.}
The authors declare no external funding or competing interests related to this work.

{
\small
\bibliographystyle{abbrvnat}
\bibliography{main.bib}

}

%%%%%%%%%%%%%%%%%%%%%%%%%%%%%%%%%%%%%%%%%%%%%%%%%%%%%%%%%%%%
\clearpage
\appendix

\section{More details on \dsname}

\begin{figure}[h]
    \centering
    \includegraphics[width=\textwidth]{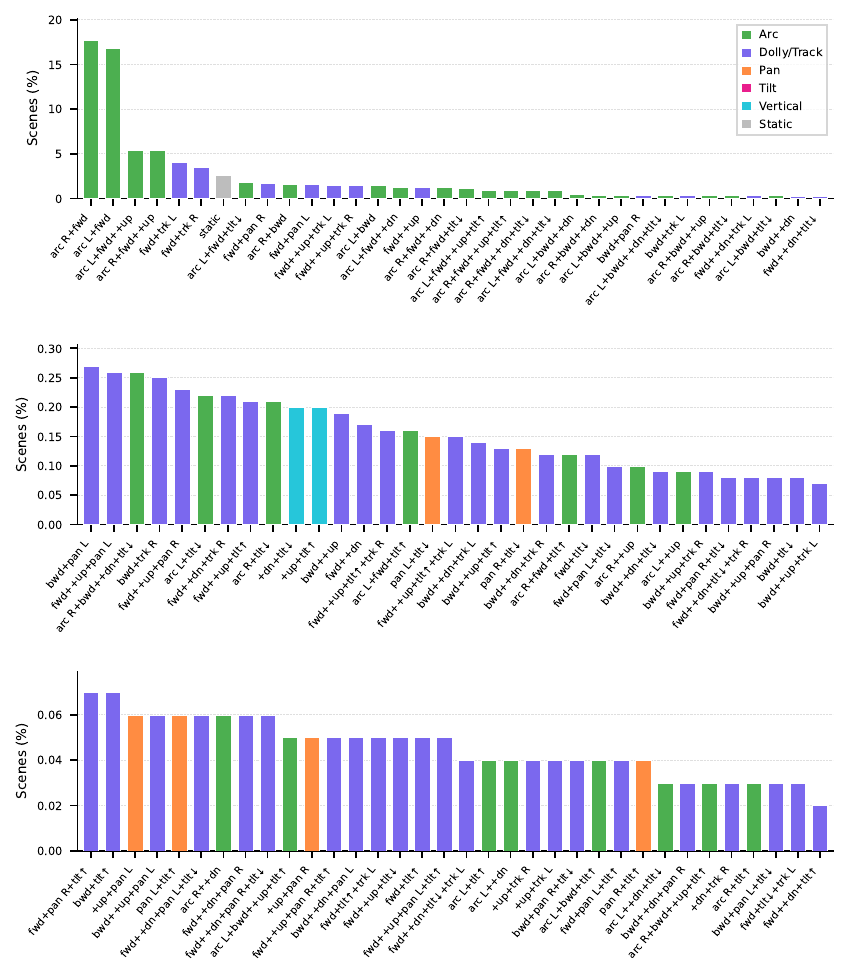}
    \caption{Complete motion category distribution of \dsname, 
    extending Figure~\ref{fig:motion_distribution} to all identified patterns. 
    Categories are sorted by frequency and color-coded by motion family.}
    \label{fig:full_motion_dist}
\end{figure}

\subsection{Full Motion Distribution}
Figure~\ref{fig:full_motion_dist} presents the complete motion category distribution of 
RealEstate10K-Cap, extending the Top-13 summary in the main text (Figure~\ref{fig:motion_distribution}) to all identified motion patterns. Each category is defined by a combination of translational axes (dolly, track, boom) and rotational axes (pan, tilt), with arcs denoting simultaneous lateral tracking and yaw and compound motions representing simultaneous multi-axis movement. The long-tail distribution—where the top two categories (arc+forward) account for $\sim$35\% of sequences while the remaining 
65\% spans 50+ diverse patterns—reflects the natural variety of real-estate footage and provides broad motion coverage.

\subsection{Data Captioning Pipeline}
\label{sec:data_pipeline}

As introduced in the main text, constructing the \dsname dataset relies on a two-stage captioning pipeline that keeps motion descriptions consistent with the trajectories while adding scene context.

\paragraph{Stage 1: Deterministic Kinematic Extraction.} We extract strict geometric motion priors directly from the relative camera poses provided by RealEstate10K, which were estimated with SLAM followed by bundle adjustment~\cite{zhou2018stereo}. We integrate local camera velocities and apply rule-based kinematic thresholds to classify translational movements (e.g., track left/right, dolly in/out, boom up/down) and rotational movements (e.g., pan, tilt, roll). Simultaneous lateral tracking and yaw are merged into compound cinematic descriptions, such as \textit{``arc left''}. This yields a rule-derived motion description of the trajectory.

\paragraph{Stage 2: VLM-based Visual Grounding.} We uniformly sample up to 32 video frames per sequence and feed them into Qwen3-VL-4B-Instruct alongside the deterministic geometric prior. Generation is performed using stochastic decoding (temperature 0.7, top-p 0.9, max new tokens 128). The prompt template is designed to strictly enforce the geometric prior while enriching it with visual context:

\vspace{0.5em}
\noindent\fbox{
    \parbox{0.96\linewidth}{
    \textit{\textbf{System Prompt:}} You act as a professional cinematographer describing a camera shot. Given the camera motion sequence reference and [n\_frames] frames from \texttt{[total\_frames]} uniformly sampled frames from the video, your task is to merge the motion sequence with the key scene elements into 1 or 2 fluid sentences.
    
    \vspace{0.5em}
    \noindent \textbf{Rules:}
    \begin{itemize}
        \setlength{\itemsep}{0pt}
        \setlength{\parskip}{0pt}
        \setlength{\parsep}{0pt}
        \item[1.] \textbf{ABSOLUTE MOTION:} You MUST use the provided motion sequence exactly as written. DO NOT alter the directions or sequence.
        \item[2.] \textbf{FORMAT:} Use `then' or `finally' to separate stages. Keep it cinematic but factual.
        \item[3.] \textbf{CONTENT:} For each stage include: movement type (pan/tilt/dolly/truck/arc/pedestal), direction, and pace (slow/medium/fast).
    \end{itemize}
    
    \noindent \textbf{Examples:}
    \begin{itemize}
        \setlength{\itemsep}{0pt}
        \setlength{\parskip}{0pt}
        \setlength{\parsep}{0pt}
        \item[-] The camera dollies forward with slight shaking, gradually closing the distance between the viewer and the fireplace.
        \item[-] The camera pans left slowly and smoothly, revealing a stone fireplace. Then it moves along the left wall, maintaining shot angle.
        \item[-] The camera dollies backward smoothly, revealing a basketball court in the foreground, then continues backward while panning slightly left.
    \end{itemize}
    
    \vspace{0.5em}
    \noindent \textbf{Reference:} \texttt{[Insert Stage 1 Prior]} \\
    \noindent \textbf{Motion description:} 
    }
}
\vspace{0.5em}

% add keyframe and trajectory, captioning illustration

\section{Detailed Training Objectives}
\label{sec:appendix_objectives}

In this section, we provide the detailed mathematical formulations for the training objectives of our proposed \name framework, complementing the high-level descriptions in the main text.

\subsection{RVQ Autoencoder Objective}
\label{sec:appendix_rvq}
To optimize the Residual Vector Quantizer (RVQ), we employ a multi-component objective tailored to the 12D kinematic properties. Let $\ell(x, \hat{x})$ denote the Smooth L1 (Huber) loss. The overall RVQ objective $\mathcal{L}_{\text{RVQ}}$ combines a global reconstruction term $\mathcal{L}_{\text{rec}} = \ell(m_t, \hat{m}_t)$ with an explicit kinematic loss $\mathcal{L}_{\text{exp}}$ and a commitment loss $\mathcal{L}_c$:
$$ \mathcal{L}_{\text{RVQ}} = \mathcal{L}_{\text{rec}} + \lambda_{\text{exp}}\mathcal{L}_{\text{exp}} + \beta\mathcal{L}_c $$
Specifically, $\mathcal{L}_{\text{exp}}$ independently supervises the translation-velocity and 6D rotation components to ensure structural integrity:
$$ \mathcal{L}_{\text{exp}} = \ell([p_t, v_t], [\hat{p}_t, \hat{v}_t]) + \ell(r_t, \hat{r}_t) + \lambda_{\text{orth}}\mathcal{L}_{\text{orth}} $$
The term $\mathcal{L}_{\text{orth}}$ enforces the orthogonality of the predicted 6D rotation vectors $\hat{r}_1, \hat{r}_2 \in \mathbb{R}^3$:
$$ \mathcal{L}_{\text{orth}} = \mathbb{E}\left[(\|\hat{r}_1\|_2 - 1)^2 + (\|\hat{r}_2\|_2 - 1)^2 + (\hat{r}_1^\top \hat{r}_2)^2\right] $$

\subsection{Masked Transformer Objective}
For the generative modeling stage, the masked transformers are optimized to predict the masked motion tokens. The base model is trained using a cross-entropy loss applied exclusively to the masked non-padding positions $\Omega$:
$$ \mathcal{L}_{\text{base}} = -\frac{1}{|\Omega|} \sum_{(b,i) \in \Omega} \log p_\theta\big(Z_{b,i}^{(0)} \mid \tilde{Z}_b^{(0)}, c_b, v_b\big) $$
Similarly, the residual masked transformer minimizes the conditional masked prediction loss over the temporal dimension for the sampled active layer $q_b$:
$$ \mathcal{L}_{\text{res}} = -\frac{1}{\sum_b S_b} \sum_b \sum_{i < S_b} \log p_\phi\big(Z_{b,i}^{(q_b)} \mid H_b^{(q_b)}, q_b, c_b, v_b\big) $$

\section{Experiment Details}

\subsection{Full List of Hyperparameters}
\label{sec:hyperparameters}

All experiments are conducted on a single NVIDIA A100 (80 GB) GPU. Table~\ref{tab:hyperparameters} lists the hyperparameters used for training both the RVQ autoencoder and the dual masked transformers.

\begin{table}[h]
    \centering
    \caption{Detailed hyperparameters for \name training.}
    \label{tab:hyperparameters}
    \begin{tabular}{llc}
        \toprule
        \textbf{Module} & \textbf{Hyperparameter} & \textbf{Value} \\
        \midrule
        \multirow{10}{*}{RVQ Autoencoder} 
        & Codebook Size ($V$) & 256 \\
        & Quantization Levels ($K$) & 4 \\
        & Quantizer Dropout Prob ($p$) & 0.2 \\
        & Kinematic Loss Weight ($\lambda_{\text{exp}}$) & 0.5 \\
        & Orthogonality Weight ($\lambda_{\text{orth}}$) & 0.1 \\
        & Commitment Weight ($\beta$) & 0.02 \\
        & Base Learning Rate & 1e-4\\
        & Learning Rate Warm-up Iterations & 500 \\
        & Batch Size & 256 \\
        & Window Size & 128 \\
        \midrule
        \multirow{7}{*}{Transformers (Base \& Residual)} 
        & Number of Layers & 4 \\
        & Number of Attention Heads & 6 \\
        & Hidden Dimension & 384 \\
        & CFG Dropout Prob ($p_{\text{drop}}$) & 0.2 \\
        & Optimizer & AdamW~\cite{loshchilov2019adamw} \\
        & Base Learning Rate & 5e-5 \\
        & Batch Size & 64 \\
        \bottomrule
    \end{tabular}
\end{table}

\subsection{Baseline Training and Evaluation Protocol}
\label{sec:baseline_protocol}

To reduce the effect of unequal training-domain exposure, we additionally fine-tune or retrain feasible baselines on RealEstate10K-Cap. In particular, CCD~\cite{jiang2024cinematographic} and E.T.~\cite{courant2024exceptional} are retrained on RealEstate10K-Cap, while GenDoP~\cite{zhang2025gendop} is evaluated in both its original and fine-tuned configurations. Director3D~\cite{li2024director3d} is evaluated using its released pretrained configuration, for which matched retraining is unavailable in our setup. All methods are evaluated using the same mixed-domain benchmark, trajectory-length protocol, and Universal CLaTr evaluator. Table~\ref{tab:main_exp_results} reports both pretrained and matched-training results where available.

\subsection{Trajectory-Length Handling}
\label{sec:trajectory_length}

\name\ requires a target sequence length to be specified at inference. For benchmark evaluation, we use the per-sample ground-truth trajectory length, denoted as $\texttt{traj\_len}$, and enforce the same target duration across all evaluated methods.

Because the VQ tokenizer temporally downsamples trajectories by a factor of four, \name\ first decodes $\lceil \texttt{traj\_len}/4 \rceil \times 4$ frames and then resamples the trajectory to $\texttt{traj\_len}$. GenDoP generates its native 120 camera poses and is resampled to the target length; CCD generates 300 native frames and is similarly resampled; Director3D's native trajectory is resampled using spherical linear interpolation (slerp) for rotations; and E.T. is run directly at the per-sample target length. We additionally audited the generated outputs used in these comparisons and found no target-length mismatches for \name, GenDoP, CCD, Director3D, and E.T.

This protocol ensures that the reported trajectory metrics are not confounded by differences in evaluated duration. The requirement of a predefined generation horizon remains a practical assumption of the current formulation. We do not use segment stitching in the reported experiments, and generation substantially beyond the training-length distribution has not been systematically evaluated.

\subsection{Controlled Dataset Comparison}
\label{sec:dataset_controlled}

To isolate the effect of training data, we train the same \name\ architecture separately on equally sized subsets ($\sim$8.5K trajectories each) of RealEstate10K-Cap and DataDoP. We keep the tokenizer, model size, optimization schedule, checkpoint-selection criterion, and evaluator fixed across the two runs. We then evaluate both models on the same four test domains.

\begin{table}[t]
  \centering
  \caption{Controlled comparison between RealEstate10K-Cap and DataDoP as training sources. The same \name\ architecture is trained on equally sized subsets of the two datasets. Lower FID and $\Delta$Div are better.}
  \label{tab:dataset_controlled}
  \footnotesize
  \setlength{\tabcolsep}{5pt}
  \resizebox{\linewidth}{!}{
  \begin{tabular}{lcccc}
    \toprule
    \textbf{Test Domain}
    & \multicolumn{2}{c}{\textbf{FID} $\downarrow$}
    & \multicolumn{2}{c}{$\Delta$\textbf{Div} $\downarrow$} \\
    \cmidrule(lr){2-3}
    \cmidrule(lr){4-5}
    & \textbf{Re10K-Cap Train}
    & \textbf{DataDoP Train}
    & \textbf{Re10K-Cap Train}
    & \textbf{DataDoP Train} \\
    \midrule
    Mixed
      & \textbf{0.967} & 1.190
      & \textbf{0.196} & 0.389 \\
    RealEstate10K
      & \textbf{1.204} & 1.458
      & \textbf{0.274} & 0.394 \\
    E.T.\ (OOD)
      & \textbf{0.956} & 1.134
      & \textbf{0.013} & 0.179 \\
    DataDoP
      & \textbf{1.431} & 1.572
      & \textbf{0.134} & 0.466 \\
    \bottomrule
  \end{tabular}}
\end{table}

As shown in Table~\ref{tab:dataset_controlled}, the RealEstate10K-Cap-trained model achieves lower FID and a smaller diversity gap across all four evaluation domains, including the DataDoP test subset itself and E.T., which is out-of-domain for both training sources. These controlled results complement the motion-distribution statistics in the main paper and provide direct evidence for the utility of RealEstate10K-Cap as a training source.

\subsection{Temporal Smoothness Analysis}
\label{sec:smoothness_analysis}

We quantitatively evaluate temporal smoothness using finite-difference translational acceleration and jerk, together with angular velocity, acceleration, and jerk. Since visual conditioning changes the evaluated subset, we compare the frame-conditioned methods against the corresponding frame-conditioned ground-truth trajectories.

Raw derivative magnitudes alone are insufficient for assessing realistic motion, as minimizing acceleration or jerk may favor over-smoothed trajectories. We therefore additionally compute the 1D Wasserstein distance between each derivative distribution and the corresponding ground-truth distribution.

\begin{table}[t]
  \centering
  \caption{Temporal smoothness analysis in the vision-conditioned setting.
  Top: mean per-frame derivative magnitudes.
  Bottom: 1D Wasserstein distance between each derivative distribution and
  ground truth (lower is closer to real camera dynamics).}
  \label{tab:smoothness}
  \footnotesize
  \setlength{\tabcolsep}{4pt}
  \resizebox{\linewidth}{!}{
  \begin{tabular}{lccccc}
    \toprule
    \textbf{Method}
    & \textbf{Trans. Acc.}
    & \textbf{Trans. Jerk}
    & \textbf{Ang. Vel.}
    & \textbf{Ang. Acc.}
    & \textbf{Ang. Jerk} \\
    \midrule
    \multicolumn{6}{c}{\textit{Mean per-frame derivative magnitude}} \\
    \midrule
    GT (reference)
      & 0.00078 & 0.00054 & 0.00460 & 0.00134 & 0.00186 \\
    \name\ (Sparse Keyframes)
      & 0.00045 & 0.00024 & 0.00745 & 0.00718 & 0.01187 \\
    \name\ (First-Frame)
      & 0.00032 & 0.00016 & 0.00629 & 0.00641 & 0.01067 \\
    GenDoP (RGBD)
      & 0.00092 & 0.00133 & 0.00394 & 0.00042 & 0.00058 \\
    \midrule
    \multicolumn{6}{c}{\textit{1D Wasserstein distance to GT} $\downarrow$} \\
    \midrule
    \name\ (Sparse Keyframes)
      & \textbf{0.00032} & \textbf{0.00029} & 0.00350 & 0.00640 & 0.01085 \\
    \name\ (First-Frame)
      & 0.00043 & 0.00036 & 0.00215 & 0.00525 & 0.00906 \\
    GenDoP (RGBD)
      & 0.00042 & 0.00069 & \textbf{0.00118} & \textbf{0.00097} & \textbf{0.00138} \\
    \bottomrule
  \end{tabular}
      \vspace{-6mm}
  }
\end{table}

Table~\ref{tab:smoothness} shows that \name\ closely matches the ground-truth translational dynamics. In particular, the sparse-keyframe setting achieves the smallest Wasserstein distances for translational acceleration and jerk. Because sparse RGB keyframes enter through soft cross-attention conditioning rather than hard-spliced motion tokens, they do not explicitly introduce trajectory discontinuities at the conditioned timestamps. In contrast, rotational acceleration and jerk remain less well matched than GenDoP-RGBD, indicating that rotational smoothness is a remaining limitation of the current model. 

\subsection{Human Validation of Universal CLaTr}
\label{sec:human_study}

We conduct a pilot blinded human study to examine whether Universal CLaTr is consistent with human judgments of text-trajectory alignment. We sample 18 benchmark captions, with six each from RealEstate10K-Cap, E.T., and DataDoP. For each caption, trajectories from five sources---\name\ (Text Only), GenDoP (finetuned), E.T. (finetuned), CCD (finetuned), and the ground-truth reference---are presented in anonymized and randomized order. 
 Three raters independently score how well each camera trajectory matches the corresponding caption on a five-point Likert scale, yielding 270 raw ratings over 90 caption--trajectory pairs.

\paragraph{Rater Instructions.}
For each caption, the five trajectories were shown as anonymized clips
(``Clip A''--``Clip E'') in randomized order. Raters were asked:

\begin{quote}\small
``How well does THIS camera motion match the caption?''
\end{quote}

The following instruction was displayed on every page:

\begin{quote}\small
``\textbf{What to rate:} for each anonymized clip, judge only how well the
camera's \underline{motion path} matches the caption. Ignore image quality
--- only caption $\leftrightarrow$ camera-motion agreement.''
\end{quote}

Each clip was rated on a five-point scale, with the displayed anchors
``5 = perfectly'', ``3 = partially'', and ``1 = unrelated''.

Inter-rater agreement is moderate, with Krippendorff's $\alpha=0.49$, while the reliability of the three-rater averaged score is higher (ICC$(2,k)=0.76$). We therefore use the mean rating across raters for the following analysis.

Universal CLaTr exhibits a positive correlation with mean human judgments across the 90 caption-trajectory pairs (Spearman $\rho=0.312$, caption-bootstrap 95\% CI $[0.089, 0.492]$; Kendall $\tau_b=0.219$). Restricting the analysis to generated trajectories yields a stronger correlation (Spearman $\rho=0.395$, 95\% CI $[0.186, 0.544]$).

At the method level, human raters assign the highest mean score to \name\ (3.30), followed by GenDoP (2.78), E.T. (2.61), and CCD (2.09). On the same evaluated subset, Universal CLaTr preserves five of the six pairwise orderings among the four generated methods. The only disagreement is between \name\ and GenDoP, whose CLaTr scores are statistically indistinguishable on this subset. These results provide preliminary evidence that Universal CLaTr reflects human assessments of text-trajectory alignment, while the limited number of captions and raters motivates larger-scale perceptual validation in future work.

\section{More \name Generation Results}

\begin{figure}
    \centering
    \includegraphics[width=\textwidth]{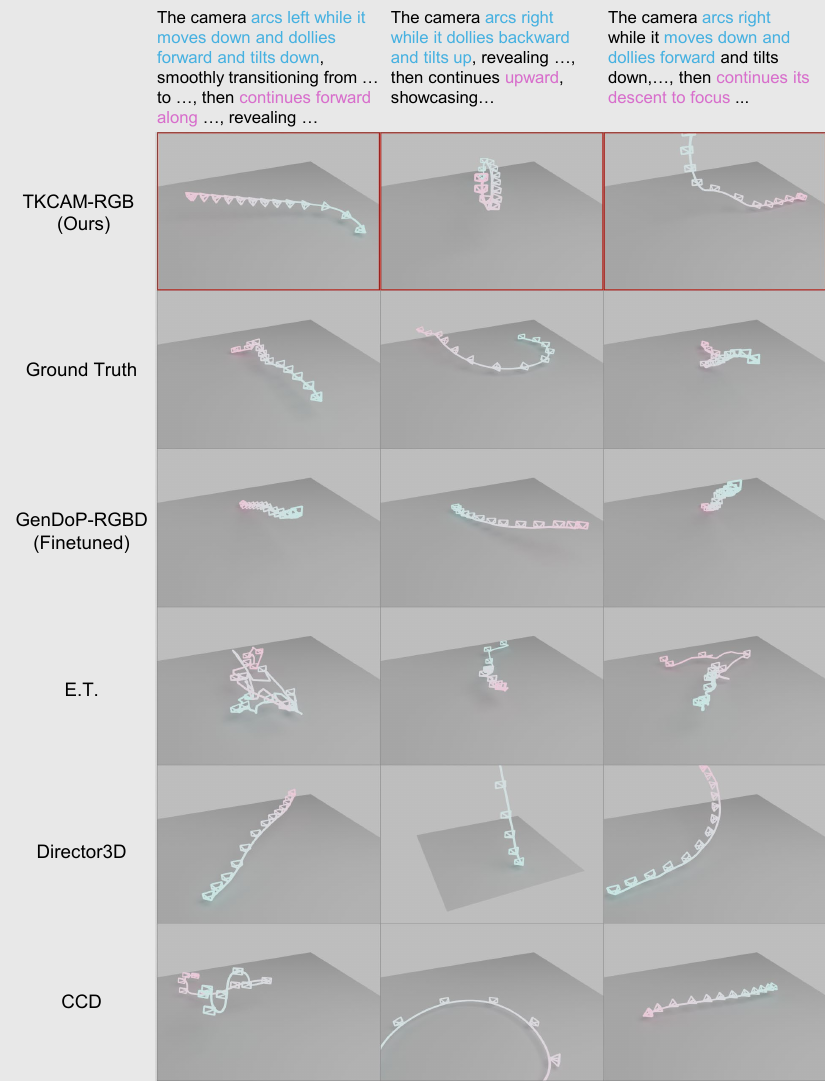}
    \caption{Additional qualitative comparisons with baselines.}
    \label{fig:more_qual}
\end{figure}

\subsection{More Qualitative Comparison}
Figure~\ref{fig:more_qual} presents additional qualitative 
comparisons across a broader range of motion types and scene contexts, supplementing 
the three examples shown in Figure~\ref{fig:baseline_comparison} of the main text. We include examples featuring predominantly translational motion (dolly-forward, lateral tracking) and complex compound motions involving simultaneous arcing, dollying, and tilting. In most of these examples, \name\ produces trajectories with correct directional semantics, while baselines exhibit the failure modes identified in the main text.

\subsection{Failure Cases of \name\ and Analysis}

\begin{figure}
    \centering
    \includegraphics[width=\textwidth]{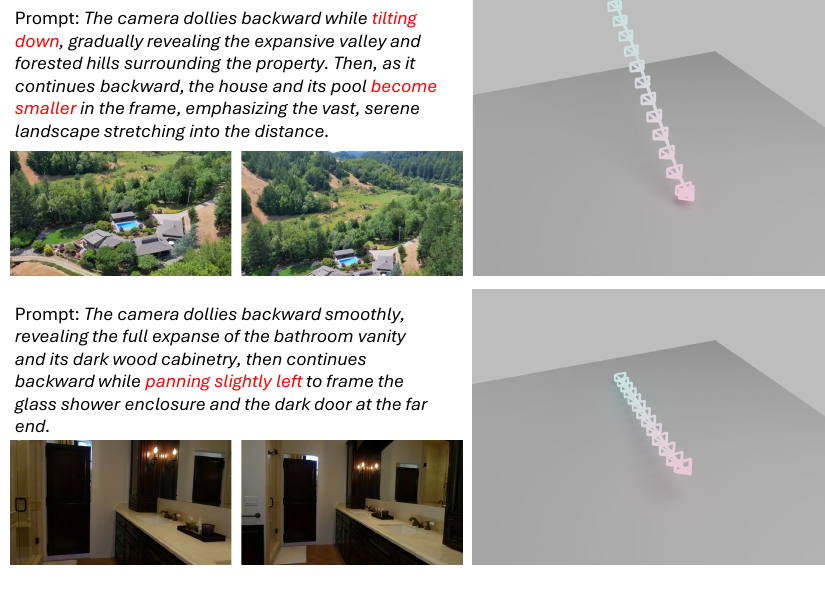}
    \caption{Two representative failure cases of \name: tilt underestimation and action omission in multi-stage sequences.}
    \label{fig:failure_cases}
\end{figure}

The vast majority of \name's generated trajectories exhibit correct
directional semantics and appropriate motion magnitude, as evidenced
by our quantitative results (Table~\ref{tab:main_exp_results}) and the qualitative comparisons in Figures~\ref{fig:baseline_comparison} and~\ref{fig:more_qual}. The following failure modes occur only in a small minority of cases, typically involving rare motion combinations or out-of-distribution prompts. Figure~\ref{fig:failure_cases} illustrates two representative failure modes of \name, with the misaligned motion keywords highlighted in \textcolor{red}{red}.

\noindent\textbf{Tilt Underestimation.}
In the first example, the prompt specifies simultaneous backward dollying and 
tilting down, yet the generated trajectory captures only the 
translational component while neglecting the vertical rotation. We attribute this 
to a distributional bias in RealEstate10K-Cap: indoor real estate tours are 
dominated by horizontal navigation, leaving tilt motions 
underrepresented in training. As a result, the model tends to under-weight 
rotational components when they co-occur with a dominant translational cue.

\noindent\textbf{Action Omission in Multi-Stage Sequences.}
In the second example, a two-stage motion (dolly backward, then 
pan slightly left) produces a trajectory that executes the 
primary dolly but omits the secondary pan. This occurs occasionally when a 
later-stage motion has low magnitude relative to the dominant motion, causing 
it to be under-sampled during masked token prediction. The issue is more 
pronounced when the secondary action is brief and semantically subtle.

Both failure modes are rare edge cases that arise specifically at the boundary of the training distribution---primarily for out-of-distribution vertical motions and low-magnitude secondary actions---rather than reflecting systematic architectural limitations. The strong quantitative performance reported in Table~\ref{tab:main_exp_results} indicates that such misalignments do not represent the typical generation behavior of \name.

\section{Ethical and Societal Impact Discussion}

\noindent\textbf{Positive Impacts.}
\name\ can make camera-path authoring more accessible by enabling 
non-expert users to generate controllable camera trajectories from natural language 
descriptions. This has direct positive applications in independent filmmaking, 
pre-visualization for film production, accessibility tools for virtual environments, 
and educational simulators for cinematography training. By open-sourcing our models 
and dataset, we further enable the research community to build upon this work for 
downstream video generation, 3D scene understanding, and embodied AI applications.

\noindent\textbf{Potential Negative Impacts.}
The ability to generate plausible, text-controlled camera trajectories could 
potentially be misused in combination with video synthesis pipelines to produce 
deceptive deepfake content or synthetic surveillance footage that appears 
cinematographically authentic. Additionally, automated cinematography generation 
may reduce demand for certain professional roles in low-budget production contexts.

\noindent\textbf{Mitigation.}
We will accompany the public release of \name with a usage policy that 
prohibits malicious or deceptive applications. As a trajectory-only model, 
\name does not generate visual content independently and requires a separate 
video synthesis component to produce any visual output, which introduces an 
additional mitigation layer. We encourage downstream users of this work to 
apply appropriate safeguards when integrating trajectory generation into 
end-to-end video generation pipelines.

%%%%%%%%%%%%%%%%%%%%%%%%%%%%%%%%%%%%%%%%%%%%%%%%%%%%%%%%%%%%

\end{document}